\documentclass[eventcountsame]{llncs}
\usepackage{paralist}
\usepackage{graphicx}
\usepackage{amssymb}
\usepackage{dsfont}
\usepackage{comment}
\usepackage{amsmath}
\usepackage{wrapfig}
\usepackage{url}

\newcounter{count_a} %
\newcounter{count_f} %
\newcounter{count_i} %
\newcounter{count_e} %
\newcounter{count_r} %
\newcommand{\nexta}{\refstepcounter{count_a}a\arabic{count_a}}
\newcommand{\nextf}{\refstepcounter{count_f}f\arabic{count_f}}
\newcommand{\nexti}{\refstepcounter{count_i}i\arabic{count_i}}
\newcommand{\nexte}{\refstepcounter{count_e}e\arabic{count_e}}
\newcommand{\nextr}{\refstepcounter{count_r}r\arabic{count_r}}

\title{Encoding Higher Level Extensions of Petri Nets in Answer Set Programming}
\titlerunning{Encoding of Higher Level Extensions of Petri Nets in ASP}

\author{Saadat Anwar\inst{1} \and Chitta Baral\inst{1} \and Katsumi Inoue\inst{2}}

\institute{SCIDSE, Arizona State University, 699 S Mill Ave, Tempe, AZ 85281, USA
	\and Principles of Informatics Research Divisions, National Institute of Informatics, Japan}

\begin{document}

\maketitle

\begin{abstract}
Answering realistic questions about biological systems and pathways similar to the ones used by text books to test understanding of students about biological systems is one of our long term research goals. Often these questions require simulation based reasoning. To answer such questions, we need formalisms to build pathway models, add extensions, simulate, and reason with them. We chose Petri Nets and Answer Set Programming (ASP) as suitable formalisms, since Petri Net models are similar to biological pathway diagrams; and ASP provides easy extension and strong reasoning abilities. We found that certain aspects of biological pathways, such as locations and substance types, cannot be represented succinctly using regular Petri Nets. As a result, we  need higher level constructs like colored tokens. In this paper, we show how Petri Nets with colored tokens can be encoded in ASP in an intuitive manner, how additional Petri Net extensions can be added by making small code changes, and how this work furthers our long term research goals. Our approach can be adapted to other domains with similar modeling needs.
\end{abstract}

\section{Introduction}
One of our long term research objectives is to develop a system that can answer questions similar to the ones given in the biological texts, used to test the understanding of the students. In order to answer such questions, we have to model pathways, add interventions / extensions to them based on the question, simulate them, and reason with the simulation results. We found Petri Nets~\cite{carl1962petri} to be a suitable formalism for modeling biological pathways, as their graphical representation is very close to the biological pathways, and they can be extended to add necessary assumptions and interventions relevant to the questions as shown in our prequel to this paper~\cite{anwar2013pniclp}. Looking through the pathways, we found that certain aspects of biological pathways, such as multiple locations and substance types (perhaps connected to these locations) cannot be represented by regular Petri Nets in a succinct manner.

\begin{figure}[htbp]
\centering
\vspace{-30pt}
\includegraphics[width=\linewidth]{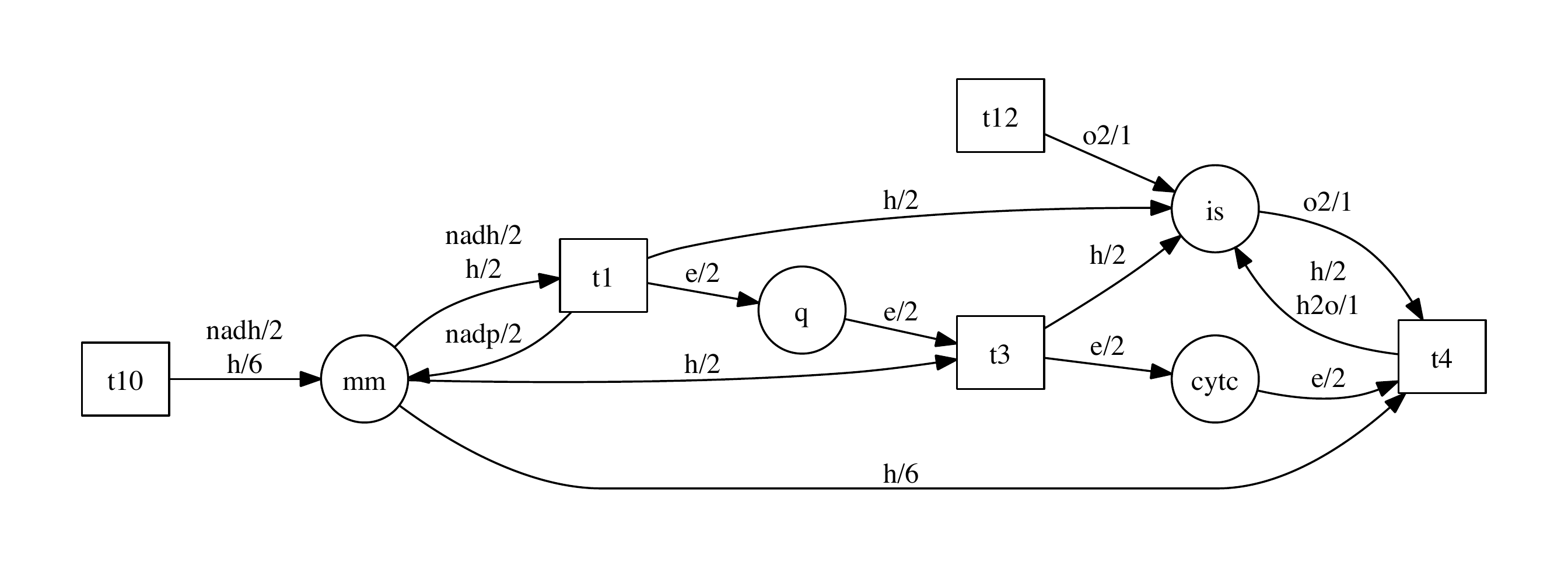}
\caption{Petri Net with tokens of colors $\{e,h,h2o,nadh,nadp,o2\}$. Circles represent places, and rectangles represent transitions. Arc weights such as ``$nadh/2,h/2$'', ``$h/2,h2o/1$'' specify the number of tokens consumed and produced during the execution of their respective transitions, where ``$nadh/2,h/2$'' means 2 tokens of color $nadh$ and 2 tokens of $h$. Similar notation is used to specify marking on places, when not present, the place is assumed to be empty of tokens.}
\label{fig:echain}
\vspace{-10pt}
\end{figure}

Consider the simplified Petri Net model of the Electron Transport Chain~\cite{CampbellBook} in Figure~\ref{fig:echain}. The chain removes high energy electrons ($e$) from NADH ($nadh$) and delivers them to Oxygen ($o2$) by using electron carriers Coenzyme Q ($q$) and Cytochrome C ($cytc$). During the process, H+ ($h$) ions are transported from the Mitochondrial Matrix ($mm$) to the Intermembrane Space ($is$). The cross-membrane H+ gradient thus produced drives ATP Synthase (not shown) to produce ATP. The transitions $t1-t4$ represent multi-protein complexes that form the chain. In order for $t1$ to fire, 2$\times$NADH and 2$\times$H+ ($nadh/2,h/2$) are required at the Mitochondrial Matrix ($mm$). It is clear that the location information embedded in this pathway is a vital part of its model. Regular Petri Nets that were a focus of our previous work~\cite{anwar2013pniclp} cannot capture this location information in a succinct way. In addition, when we use place nodes to represent locations (as in Figure~\ref{fig:echain}), regular (uncolored) tokens do not provide sufficient fidelity to represent various token types needed as input to a transition. As a result, we have to use Petri Nets with colored tokens~\cite{peterson1980note}  to model such biological pathways\footnote{Though a Petri Net with colored tokens can be converted into a regular Petri Net without colored tokens, they are usually too large and cumbersome, hence inconvenient.}. In contrast to our previous work, the place nodes in this Petri Net model represent locations rather than substances. Even electron-carrier (substances) $q,cytc$ are locations for electrons ($e$) to be stored and shuttled. Colored tokens also provide a mechanism for differentiating between separate quantities of the same substance present in multiple locations (a common occurrence in biological systems).

Numerous Petri Net modeling and simulation systems exist~\cite{jensen2007coloured,SnoopyPN,nagasaki2010cell,kummer1999renew}, but we did not find them to be suitable for our application, either due to limited adaptability outside their intended application domain, limited extendibility, or ease of extendibility. In addition, most systems did not explore all possible state evolutions, allowed different firing semantics, or provided a way to guide the search through specification of partial state as {\em way-points}. We found the features, such as intuitive encoding, easy extendibility, and strong reasoning capability in Answer Set Programming (ASP), which is a declarative programming language with numerous competitive solvers~\cite{gebser2007first}. It has been effectively used in various domains, such as spacecrafts, work flows, natural language processing, and biological systems~\cite{brewka2011answer}. The suitability of ASP to analyze Petri Nets is further reinforced over other techniques, such as process algebra, temporal logics, and mathematical equations when one considers the restrictions on Petri Nets imposed by mathematical techniques~\cite{murata1989petri}, the cumbersomeness of encoding in $\pi$-calculus even for small models~\cite{van2005pi}, or the lack of applicability to higher level Petri Net extensions.

Previous work on Petri Net to ASP translation has been limited to specific classes of Petri Nets, such as regular Petri Nets
~\cite{heljanko2001bounded} and Simple Logic Petri Nets (SLPN)~\cite{LogicPetriNets}, focusing on analyzing their properties. Neither used colored tokens. Please see our previous work~\cite{anwar2013pniclp} for more details. Though our focus in the current work is on biological questions, our approach is equally suited for hypothesis verification during drug design, drug interaction and biological systems model development. It can also be applied to other domains where Petri Nets are used for modeling and simulation, such as work flows, embedded systems, and industrial control.

Thus, the main contributions of this paper are as follows. In Section~\ref{sec:enc_cpn} we show how ASP allows intuitive declarative encoding of higher level Petri Net extension of colored tokens~\cite{peterson1980note}. We then show how additional extensions can be incorporated in our encoding by making small changes. In this regard, we present changing the firing semantics (Section~\ref{sec:enc_max_firing}), priority transitions (Section~\ref{sec:enc_priority}), and timed transitions (Section~\ref{sec:enc_dur}). We show how Petri Nets and our encoding fit into our ultimate research goals of answering questions about biological pathways. We start with a brief background on ASP, multisets, and Petri Nets.

\section{Fundamentals}
{\bf Answer Set Programming (ASP)} is a declarative logic programming language based on the Stable Model Semantics~\cite{StableModels}. Code presented in this paper follows the Clingo~\cite{Clingo} syntax.
The reader is referred to~\cite{Clingo,Baral2003} for the syntax and semantics of Answer Set Programs.

A {\bf multiset} $A$ over a domain set $D$ is a pair $\langle D,m \rangle$, where $m: D \rightarrow \mathds{N}$ is a function giving the multiplicity of $d \in D$ in $A$. Given two multsets $A = \langle D,m_A \rangle, B = \langle D,m_B \rangle$, $A \odot B$ if $\forall d \in D: m_A(d) \odot m_B(d)$, where $\odot \in \{<,>,\leq,\geq,=\}$, and $A \neq B$ if $\exists d \in D : m_A(d) \neq m_B(d)$. Multiset sum/difference is defined in the usual way.  We use the short-hands $d \in A$ to represent $m_A(d) > 0$, $A = \emptyset$ to represent $\forall d \in D, m(d) = 0$, $A \otimes n$ to represent $\forall d \in D, m(d) \otimes n$, where $n \in \mathds{N}$, $\otimes \in \{<,>,\leq,\geq,=,\neq\}$. We use the notation $d/n \in A$ to represent that $d$ appears $n$-times in $A$; we drop $A$ when clear from context. The reader is referred to \cite{syropoulos2001mathematics} for details.

A basic {\bf Petri Net}~\cite{carl1962petri} is a bipartite graph of a finite set of place nodes $P=\{p_1,\dots,p_n\}$, and transition nodes $T=\{t_1,\dots,t_m\}$ connected through directed arcs $E=E^+ \cup E^-$. An arc goes from a place to a transition $E^- \subseteq P \times T$ or a transition to a place $E^+ \subseteq T \times P$. The state of a Petri Net is defined by the token allocation of all place nodes, collectively called its \textit{marking} $M=(M(p_1),\dots,M(p_n)), M(p_i) \in \mathds{N}$. Arc weights $W : E \rightarrow \mathds{N} \setminus \{0\}$ specify the number of tokens consumed or produced from place nodes at the head or tail of the arcs due to firing of a transition. Modeling capability of basic Petri Nets is enhanced by adding reset, inhibit and read arcs. Reset arcs $R: T \rightarrow 2^P$ remove all tokens from their source places when fired. Inhibitor arcs $I : T \rightarrow 2^P$ prevent their transitions from firing until their source places are empty. Read arcs $Q \subseteq P \times T$ prevent their transitions from firing until their source places have at least the tokens specified by read arc weights $QW: Q \rightarrow \mathds{N} \setminus \{0\}$.

Higher level Petri Nets extend the notion of tokens to typed (or colored) tokens. 
A {\bf Petri Net with Colored Tokens} (with reset, inhibit and read arcs) is a tuple $PN^C=(P,T,E,C,W,R,I,Q,QW)$, where $P,T,E,R,I,Q$ are the same as for basic Petri Nets, $C=\{c_1,\dots,c_l\}$ is a finite set of colors (or types), and arc weights $W : E \rightarrow \langle C,m \rangle$, $QW : Q \rightarrow \langle C,m \rangle$ are specified as multi-sets of colored tokens over color set $C$. The state (or marking) of place nodes $M(p_i) = \langle C,m \rangle, p_i \in P$ is specified as a multiset of colored tokens over set $C$.

We will now define a number of concepts about Petri Nets used in this paper. The {\bf initial marking} is the initial token assignment of place nodes and is represented by $M_0$. The marking at time-step $k$ is written as $M_k$. The {\bf pre-set} (or input-set) of a transition $t$ is $\bullet t = \{ p \in P | (p,t) \in E^- \}$, while the {\bf post-set} (or output-set) is $t \bullet = \{ p \in P | (t,p) \in E^+ \}$. A transition $t$ is {\bf enabled} with respect to marking $M$, $enabled_M(t)$, if each of its input places $p$ has at least the number of colored-tokens as the arc-weight $W(p,t)$\footnote{In the following text, for simplicity, we will use $W(p,t)$ to mean $W(\langle p,t \rangle)$. We use similar simpler notation for $QW$.}, each of its inhibiting places $p_i \in I(t)$ have zero tokens and each of its read places $p_q : (p_q,t) \in Q$ have at least the number of colored-tokens as the read-arc-weight $QW(p_q,t)$, i.e. ($\forall p \in \bullet t, W(p,t) \leq M(p)) \wedge (\forall p \in I(t), M(p) = \emptyset) \wedge (\forall (p,t) \in Q, M(p) \geq QW(p,t))$ for a given $t$. Any number of enabled transitions may fire simultaneously as long as they don't conflict. The set $T_k=\{t_{k_1},\dots,t_{k_n}\} \subseteq T$ of such simultaneously firing transitions is called a {\bf firing set}. Execution of a firing set $T_k$ on a marking $M_k$ computes a new marking $M_{k+1}$ as: 
$\forall p \in P \setminus R(T_k), M_{k+1}(p) = M_k(p) -  \sum_{\substack{t \in T_k  \wedge p \in \bullet t}} W(p,t) + \sum_{\substack{t \in T_k  \wedge p \in t \bullet}} W(t,p)$, 
$\forall p \in R(T_k), M_{k+1}(p) = \sum_{\substack{t \in T_k \wedge p \in t \bullet}} W(t,p)$, 
where $R(T_k)=\bigcup_{t \in T_k} R(t)$.
A set of transitions $T_c \subseteq \{ t : enabled_{M_k}(t) \}$ is in conflict {\bf conflict} in $PN^C$ with respect to $M_k$ if firing them will consume more tokens than are available at one of their common input places, i.e., 
$\exists p \in P :
M_k(p) < (\sum_{t \in T_c \wedge p \in \bullet t}{W(p,t)} + \sum_{\substack{t \in T_c  \wedge p \in R(t)}}{M_k(p)})$\footnote{The reset arc is involved here because we use a modified execution semantics of reset arcs compared to the standard definition~\cite{araki1976some}. Even though both capture similar operation, our definition allows us to model elimination of all quantity of a substance as soon as it is produced, even in a maximal firing set semantics. Our semantics considers reset arc's token consumption in contention with other arcs, while the standard definition does not.}. 
An {\bf execution sequence} is the simulation of a firing sequence $\sigma = T_0,T_1,\dots,T_k$, where each $T_i \subseteq T, 0 \leq i \leq k$ is a firing set. It is the transitive closure of executions, where subsequent markings become the initial marking for the next transition set. Thus in the execution sequence $X = M_0, T_0, M_1, T_1, \dots, T_k, M_{k+1}$, the firing of $T_0$ at $M_0$ produces $M_1$, which becomes initial marking for $T_1$.

If the Figure~\ref{fig:echain}  Petri Net has the marking: $M_0(mm)=[nadh/2,h/6]$, $M_0(q)=[e/2]$, $M_0(cytc)=[e/2]$, $M_0(is)=[o2/1]$, then transitions $t1,t3,t4$ are enabled. However, either $\{t1,t3\}$ or $\{t4\}$ can fire simultaneously in a single firing at time 0 due to limited $h$ tokens in $mm$. $t4$ is said to be in conflict with $t1,t3$.

\section{Translating Petri Nets with Colored Tokens to ASP}\label{sec:enc_cpn}
In this section we present an ASP encoding of a Petri Net with Colored Tokens $PN^C$, with an initial marking $M_0$ and a simulation length $k$. This work extends our encoding of regular Petri Nets in \cite{anwar2013pniclp}. The following sections will show how Petri Net extensions can be easily added to this initial encoding.
We represent the Petri Net $PN^C$ with initial marking $M_0$, and simulation time with the following facts and rules.
\begin{description}
\item[\nextf:\label{f:place}] Facts \texttt{\small place($p_i$)} where $p_i \in P$ is a place. 
\item[\nextf:\label{f:trans}] Facts \texttt{\small trans($t_j$)} where $t_j \in T$ is a transition.
\item[\nextf:\label{f:col}] Facts \texttt{\small col($c_k$)} where $c_k \in C$ is a color. %
\item[\nextf:\label{f:ptarc}] Rules \texttt{\small ptarc($p_i,t_j,n_c,c,ts_k$) :- time($ts_k$).} for each $(p_i,t_j) \in E^-$, $c \in C$, $n_c=m_{W(p_i,t_j)}(c) : n_c > 0$.\footnote{The time parameter $ts_k$ allows us to capture reset arcs, which consume tokens equal to the current (time-step based) marking of their source nodes.} %
\item[\nextf:\label{f:tparc}] Rules \texttt{\small tparc($t_i,p_j,n_c,c,ts_k$) :- time($ts_k$).} for each $(t_i,p_j) \in E^+$ , $c \in C$, $n_c=m_{W(t_i,p_j)}(c) : n_c > 0$. %
\item[\nextf:\label{f:rptarc}] Rules \texttt{\small ptarc($p_i,t_j,n_c,c,ts_k$) :- holds($p_i,n_c,c,ts_k$),  num($n_c$), $n_c$>0, time($ts_k$).}  for each $(p_i,t_j):$ $p_i \in R(t_j)$, $c \in C$, $n_c=m_{M_k(p_i)}(c)$. %
\item[\nextf:\label{f:iptarc}] Rules \texttt{\small iptarc($p_i,t_j,1,c,ts_k$) :- time($ts_k$).} for each $(p_i,t_j): p_i \in I(t_j)$, $c \in C$. %
\item[\nextf:\label{f:tptarc}] Rules \texttt{\small tptarc($p_i,t_j,n_c,c,ts_k$) :- time($ts_k$).} for each $(p_i,t_j) \in Q$, $c \in C$, $n_c=m_{QW(p_i,t_j)}(c) : n_c > 0 $. %
\item[\nexti:\label{i:init}] Facts \texttt{\small holds($p_i,n_c,c,0$).} for each place $p_i \in P, c \in C, n_c=m_{M_0(p_i)}(c)$.
\item[\nextf:\label{f:time}] Facts \texttt{\small time($ts_i$)} where $0 \leq ts_i \leq k$ are the discrete simulation time steps. %
\item[\nextf:\label{f:num}] Facts \texttt{\small num($n$)} where $0 \leq n \leq ntok$ are token quantities\footnote{The token count predicate \texttt{num}'s limit can be arbitrarily selected to be higher than the expected token count. It is there for efficient ASP grounding and not to impose a limit on the token quantity.}
\end{description}

Next, we encode Petri Net's {\bf execution behavior}, which proceeds in discrete time steps.
For a transition $t_i$ to be enabled, it must satisfy the following conditions:
\begin{inparaenum}[(i)]
\item $\nexists p_j \in \bullet t_i : M(p_j) < E^-(p_j,t_i)$,
\item $\nexists p_j \in I(t_i) : M(p_j) > 0$, and
\item $\nexists (p_j,t_i) \in Q :  M(p_j) < QW(p_j,t_i)$
\end{inparaenum}.
These three conditions are encoded as $e\ref{e:ne:ptarc},e\ref{e:ne:iptarc},e\ref{e:ne:tptarc}$, respectively and we encode the absence of any of these conditions for a transition as $e\ref{e:enabled}$:
\begin{description}
\item[\nexte:\label{e:ne:ptarc}] \texttt{\small notenabled(T,TS) :- ptarc(P,T, N,C,TS), holds(P,Q,C,TS),  
   place(P),  \\ trans(T), time(TS), num(N), num(Q), col(C), Q<N.} %

\item[\nexte:\label{e:ne:iptarc}] \texttt{\small notenabled(T,TS) :- iptarc(P,T,N,C,TS), holds(P,Q,C,TS), 
   place(P), \\ trans(T), time(TS), num(N), num(Q), col(C), Q>=N.} %

\item[\nexte:\label{e:ne:tptarc}] \texttt{\small notenabled(T,TS) :- tptarc(P,T,N,C,TS), holds(P,Q,C,TS),
    place(P), \\ trans(T), time(TS), num(N), num(Q), col(C), Q<N.} %

\item[\nexte:\label{e:enabled}] \texttt{\small enabled(T,TS) :- trans(T), time(TS), not notenabled(T,TS).} %

\end{description}
Rule $e\ref{e:ne:ptarc}$ captures the existence of an input place $P$ with insufficient number of tokens for transition $T$ to fire. Rule $e\ref{e:ne:iptarc}$ captures existence of a non-empty source place $P$ of an inhibitor arc to $T$ preventing $T$ from firing. Rule $e\ref{e:ne:tptarc}$ captures existence of a source place $P$ with less than arc-weight tokens required by the read arc to transition $T$ for $T$ to be enabled. The, \texttt{\small holds(P,Q,C,TS)} predicate captures the marking of place $P$ at time $TS$ as $Q$ tokens of color $C$. Rule $e\ref{e:enabled}$ captures enabling of transition $T$ when no reason for it to be not enabled is determined by $e\ref{e:ne:ptarc},e\ref{e:ne:iptarc},e\ref{e:ne:tptarc}$. In a biological context, this enabling is equivalent to a reaction's pre-conditions being satisfied. A reaction can proceed when its input substances are available in the required quantities, it is not inhibited, and any required activation quantity of activating substances is available.

Any subset of enabled transitions can fire simultaneously at a given time-step. We select a subset of fireable transitions using a choice rule:
\begin{description}
\item[\nexta:\label{a:fires}] \texttt{\small \{fires(T,TS)\} :- enabled(T,TS), trans(T), time(TS).} %
\end{description} 

The choice rule $a\ref{a:fires}$ either picks an enabled transition $T$ for firing at time $TS$ or not. The combined effect over all transitions is to pick a subset of enabled transitions to fire. Whether these transitions are in conflict are checked by later rules $a\ref{a:overc:place},a\ref{a:overc:gen},a\ref{a:overc:elim}$. In a biological context, the multiple firing models parallel processes occurring simultaneously. The marking is updated according to the firing set using the following rules:
\begin{description}
\item[\nextr:\label{r:add}] \texttt{\small add(P,Q,T,C,TS) :- fires(T,TS), tparc(T,P,Q,C,TS), time(TS).} %
\item[\nextr:\label{r:del}] \texttt{\small del(P,Q,T,C,TS) :- fires(T,TS), ptarc(P,T,Q,C,TS), time(TS). } %

\item[\nextr:\label{r:totincr}] \texttt{\small tot\_incr(P,QQ,C,TS) :- col(C), 
   QQ = \#sum[add(P,Q,T,C,TS) = Q : num(Q) : trans(T)], 
   time(TS), num(QQ), place(P).} %

\item[\nextr:\label{r:totdecr}] \texttt{\small tot\_decr(P,QQ,C,TS) :- col(C),
   QQ = \#sum[del(P,Q,T,C,TS) = Q : num(Q) : trans(T)], 
   time(TS), num(QQ), place(P).} %

\item[\nextr:\label{r:nextstate}] \texttt{\small holds(P,Q,C,TS+1):-place(P),num(Q;Q1;Q2;Q3),time(TS),time(TS+1),col(C),
  holds(P,Q1,C,TS), tot\_incr(P,Q2,C,TS), 
    tot\_decr(P,Q3,C,TS), Q=Q1+Q2-Q3.} %
\end{description}

Rules $r\ref{r:add}$ and $r\ref{r:del}$ capture that $Q$ tokens of color $C$ will be added or removed to/from place $P$ due to firing of transition $T$ at the respective time-step $TS$. Rules $r\ref{r:totincr}$ and $r\ref{r:totdecr}$ aggregate these tokens for each $C$ for each place $P$ (using aggregate assignment \texttt{\small QQ = \#sum[\dots ]}) at the respective time-step $TS$. Rule $r\ref{r:nextstate}$ uses the aggregates to compute the next marking of $P$ for color $C$ at the time-step ($TS+1$) by subtracting removed tokens and adding added tokens to the current marking. In a biological context, this captures the effect of a process / reaction, which consumes its inputs and produces outputs for the downstream processes. We capture token  overconsumption using the following rules:
\begin{description}
\item[\nexta:\label{a:overc:place}] \texttt{\small consumesmore(P,TS) :- holds(P,Q,C,TS), tot\_decr(P,Q1,C,TS), Q1 > Q. } %
\item[\nexta:\label{a:overc:gen}] \texttt{\small consumesmore :- consumesmore(P,TS).} %
\item[\nexta:\label{a:overc:elim}] \texttt{\small :- consumesmore.} %
\end{description} 

Rule $a\ref{a:overc:place}$ determines whether firing set selected by $a\ref{a:fires}$ will cause overconsumption of tokens at $P$ at time $TS$ by comparing available tokens to aggregate tokens removed as determined by $r\ref{r:totdecr}$. Rule $a\ref{a:overc:gen}$ generalizes the notion of overconsumption, while rule $a\ref{a:overc:elim}$ eliminates answer with such overconsumption. In a biological context, conflict (through overconsumption) models the limitation of input substances, which dictate which downstream processes can occur simultaneously.

\begin{proposition}
Let $PN^C$ be a Petri Net with colored tokens, reset, inhibit, and read arcs and $M_0$ be an initial marking and let $\Pi^3(PN^C,M_0,k)$ be the ASP encoding of $PN^C$ and $M_0$ over a simulation of length $k$ as defined in Section \ref{sec:enc_cpn}. Then $X^3=M_0,T_0,M_1,\dots,T_k$  is an execution sequence of $PN^C$ (with respect to $M_0$) iff there is an answer-set A of $\Pi^3(PN^C,M_0,k)$ such that:
$\{ fires(t,j) : t \in T_j, 0 \leq j \leq k \} = \{ fires(t,ts) : fires(t,ts) \in A \}$
and
$\{ holds(p,q,c,j) : p \in P, c/q  \in M_j(p), 0 \leq j \leq k \} = \{ holds(p,q,c,ts) : holds(p,q,c,ts) \in A\}$
\end{proposition}

\subsection{Example Execution}\label{sec:example_exec}
Given the above translation rules, the following facts and rules encode the Petri Net in Figure~\ref{fig:echain} with an initial marking of zero tokens\footnote{We show a few of the \texttt{\small tparc/5, ptarc/5, holds/4} to illustrate the approach, the rest of them can be encoded in a similar fashion.}:
{\small
\begin{verbatim}
time(0..5). num(0..30). place(mm;is;q;cytc). trans(t1;t3;t4;t10;t12). 
col(nadh;h;e;nadp;h2o;o2). holds(mm,0,nadh,0). holds(mm,0,h,0).
tparc(t12,is,1,o2,TS):-time(TS). tparc(t10,mm,6,h,TS):-time(TS).   
tparc(t10,mm,2,nadh,TS):-time(TS). ptarc(mm,t1,2,nadh,TS):-time(TS). 
\end{verbatim}
}
We get thousands of answer-sets, for example\footnote{We are only showing colored tokens with non-zero quantity. Also, we are representing \texttt{fires(t1,ts),\dots,fires(tm,ts)} as \texttt{fires(t1;\dots;tm,ts)} to conserve space.}:
{\small
\begin{verbatim}
fires(t10;t12,0) holds(is,1,o2,1) holds(mm,6,h,1) holds(mm,2,nadh,1)
fires(t1;t10;t12,1) holds(is,2,h,2) holds(is,2,o2,2) holds(mm,10,h,2) 
   holds(mm,2,nadh,2) holds(mm,2,nadp,2) holds(q,2,e,2)
fires(t1;t3;t10;t12,2) holds(cytc,2,e,3) holds(is,6,h,3) holds(is,3,o2,3) 
   holds(mm,12,h,3) holds(mm,2,nadh,3) holds(mm,4,nadp,3) holds(q,2,e,3)
fires(t1;t3;t4;t10;t12,3) holds(cytc,2,e,4) holds(is,12,h,4) 
   holds(is,1,h2o,4) holds(is,3,o2,4) holds(mm,8,h,4) holds(mm,2,nadh,4) 
   holds(mm,6,nadp,4) holds(q,2,e,4)
fires(t3;t4;t10;t12,4) holds(cytc,2,e,5) holds(is,16,h,5) 
   holds(is,2,h2o,5) holds(is,3,o2,5) holds(mm,6,h,5) holds(mm,4,nadh,5) 
   holds(mm,6,nadp,5) fires(t1;t10;t12,5)
\end{verbatim}
}

\subsection{Changing Firing Semantics}\label{sec:enc_max_firing}
The above code implements a \textit{set firing} semantics, which can produce a large number of answer-sets\footnote{A subset of a firing set can also be fired as a firing set by itself.}. In biological domain, it is often preferable to simulate the maximum parallel activity at each time step. We accomplish this by enforcing the \textbf{maximal firing set} semantics by extending its encoding for regular Petri Nets in \cite{anwar2013pniclp} to colored tokens as follows:
\begin{description}
\item[\nexta:\label{a:maxfire:cnh}] \texttt{\small could\_not\_have(T,TS):-enabled(T,TS),not fires(T,TS),
   ptarc(S,T,Q,C,TS), holds(S,QQ,C,TS), tot\_decr(S,QQQ,C,TS),
   Q > QQ - QQQ.} %
\item[\nexta:\label{a:maxfire:elim}] \texttt{\small :- not could\_not\_have(T,TS), time(TS), enabled(T,TS), not fires(T,TS), trans(T).}
\end{description} %
Rule $a\ref{a:maxfire:cnh}$ captures the fact that transition $T$, though enabled, could not have fired at $TS$, as its firing would have caused overconsumption. Rule $a\ref{a:maxfire:elim}$ eliminates any answers where an enabled transition could have fired without causing overconsumption but did not. This modification reduces the number of answers produced for the Petri Net in Figure~\ref{fig:echain} to 4. We can encode other firing semantics with similar ease\footnote{For example, if \textit{interleaved} semantics is desired, rules $a\ref{a:maxfire:cnh},a\ref{a:maxfire:elim}$ can changed to capture and eliminate answer-sets in which more than one transition fires in a firing set as:
\vspace{-5pt}
\begin{description}
\item[a\ref{a:maxfire:cnh}':] \texttt{\small more\_than\_one\_fires :- fires(T1,TS),fires(T2,TS),T1!=T2,time(TS).}
\item[a\ref{a:maxfire:elim}':] \texttt{\small :-more\_than\_one\_fires.}
\end{description}}.
We now look at how additional extensions can be easily encoded by making small code changes.

\section{Extension - Priority Transitions}\label{sec:enc_priority}
Priority transitions enable ordering of Petri Net transitions, favoring high priority transitions over lower priority ones~\cite{best1992petri}. In a biological context, this is used to model primary (or dominant) vs. secondary pathways / processes in a biological system. This prioritization may be due to an intervention (such as prioritizing elimination of a metabolite over recycling it).

A {\bf Priority Colored Petri Net} with reset, inhibit, and read arcs is a tuple $PN^{pri} = (P,T,E,C,W,R,I,Q,QW,Z)$, where:
$P,T,E,C,W,R,I,Q,QW$ are the same as for $PN^C$, and 
$Z : T \rightarrow \mathds{N}$ is a priority function that assigns priorities to transitions. Lower number signifies higher priority.
A {\bf transition $t_i$ is enabled in $PN^{pri}$} if it would be enabled in $PN^C$ (with respect to M) and there isn't another transition $t_j$ that would be enabled in $PN^C$ (with respect to M) s.t. $Z(t_j) < Z(t_i)$.
We add the following facts and rules to encode transition priority and enabled priority transitions:
\begin{description}
\item[\nextf:\label{f:pr}] Facts \texttt{\small transpr($t_i$,$pr_i$)} where $pr_i$ is $t_i's$ priority. %
\item[\nexta:\label{a:prne}] \texttt{\small notprenabled(T,TS) :- enabled(T,TS), transpr(T,P), 
   enabled(TT,TS), \\ transpr(TT,PP), PP < P.} %
\item[\nexta:\label{a:prenabled}] \texttt{\small prenabled(T,TS) :- enabled(T,TS), not notprenabled(T,TS).} %
\end{description}

Rule $a\ref{a:prne}$ captures that an enabled transition $T$ is not priority-enabled, if there is another enabled transition with higher priority at $TS$. Rule $a\ref{a:prenabled}$ captures that transition $T$ is priority-enabled at $TS$ since there is no enabled transition with higher priority. We replace rules $a\ref{a:fires},a\ref{a:maxfire:cnh},a\ref{a:maxfire:elim}$ with $a\ref{a:prfires},a\ref{a:prmaxfire:cnh},a\ref{a:prmaxfire:elim}$ respectively to propagate priority as follows:

\begin{description}
\item[\nexta:\label{a:prfires}] \texttt{\small \{fires(T,TS)\} :- prenabled(T,TS), trans(T), time(TS).} %

\item[\nexta:\label{a:prmaxfire:cnh}] \texttt{\small could\_not\_have(T,TS) :- prenabled(TS,TS), not fires(T,TS), 
   \\ ptarc(S,T,Q,C,TS), holds(S,QQ,C,TS), tot\_decr(S,QQQ,C,TS),
   Q > QQ - QQQ.} %

\item[\nexta:\label{a:prmaxfire:elim}] \texttt{\small :- prenabled(tr,TS), not fires(tr,TS), time(TS).} %
\end{description}

Rules $a\ref{a:prfires},a\ref{a:prmaxfire:cnh},a\ref{a:prmaxfire:elim}$ perform the same function as $a\ref{a:fires},a\ref{a:maxfire:cnh},a\ref{a:maxfire:elim}$, except that they consider only priority-enabled transitions as compared all enabled transitions.

\begin{proposition}
Let $PN^{pri}$ be a Petri Net with colored tokens, reset, inhibit, read arcs and priority based transitions and $M_0$ be an initial marking and let $\Pi^5(PN^{pri},M_0,k)$ be the ASP encoding of $PN^{pri}$ and $M_0$ over a simulation of length $k$ as defined in Section \ref{sec:enc_priority}. Then $X^5=M_0,T_0,M_1,\dots,T_k$  is an execution sequence of $PN^{pri}$ (with respect to $M_0$) iff there is an answer-set A of $\Pi^5(PN^{pri},M_0,k)$ such that:
$\{ fires(t,j) : t \in T_j, 0 \leq j \leq k \}  = \{ fires(t,ts) : fires(t,ts) \in A \}$, and
$\{ holds(p,q,c,j) : p \in P, c/q \in M_j(p), 0 \leq j \leq k \} \\ = \{ holds(p,q,c,ts) : holds(p,q,c,ts) \in A\}$
\end{proposition}

\section{Extension - Timed Transitions}\label{sec:enc_dur}
Biological processes vary in time required for them to complete. Timed transitions~\cite{ramchandani1974analysis} model this variation of duration. The timed transitions can be reentrant or non-reentrant\footnote{A {\bf reentrant} transition is like a vehicle assembly line, which accepts new parts while working on multiple vehicles at various stages of completion; whereas a {\bf non-reentrant} transition only accepts new input when the current processing is finished.}. We extend our encoding to allow reentrant timed transitions.

\begin{figure}[htbp]
\centering
\vspace{-20pt}
\includegraphics[width=\linewidth]{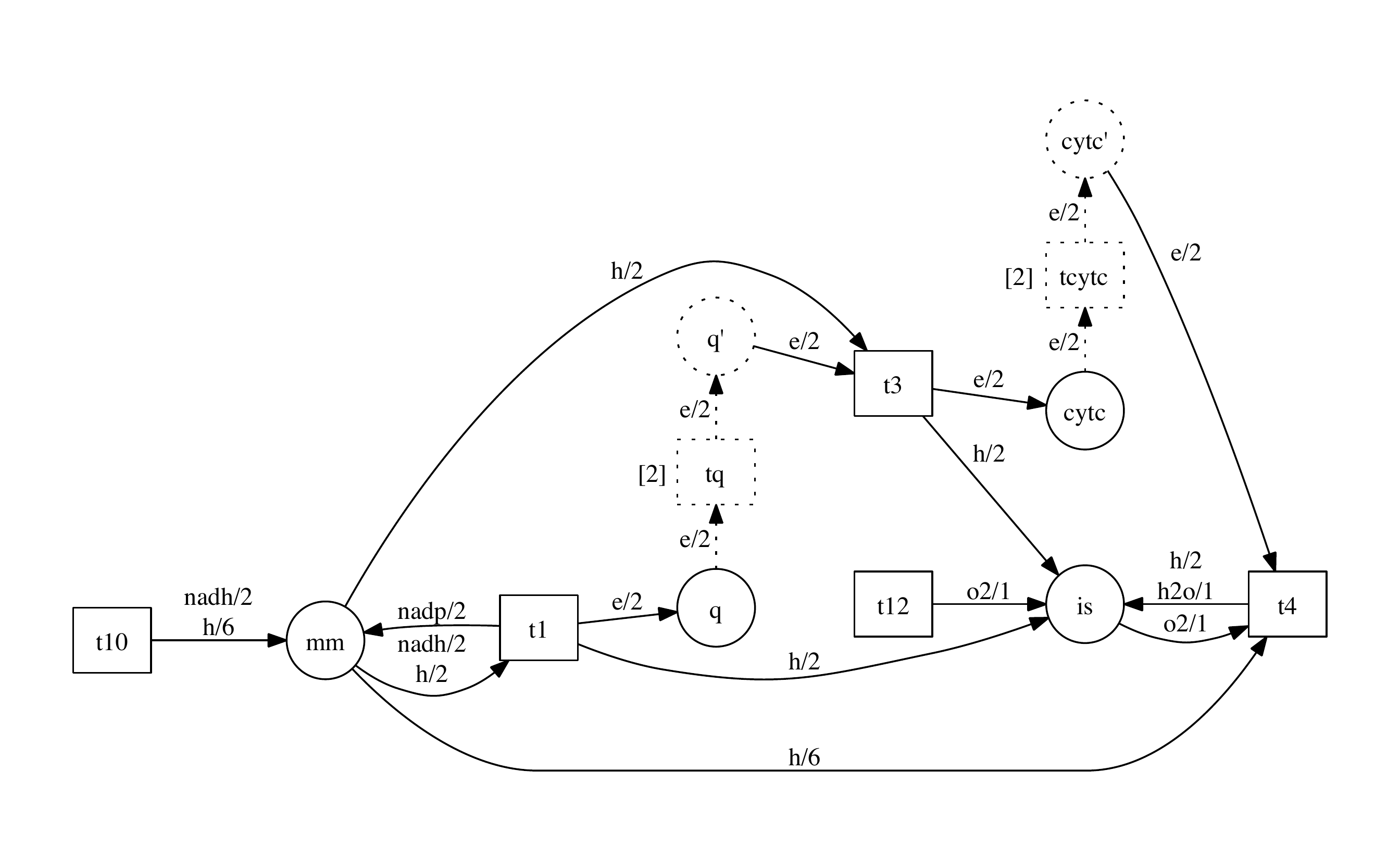}
\caption{An extended version of the Petri Net model from Fig.~\ref{fig:echain}. The new transitions $tq,tcytc$ have a duration of 2 each (shown in square brackets (``[ ]'') next to the transition). When missing, transition duration is assumed to be 1.}
\label{fig:echaintm}
\end{figure}

A {\bf Priority Colored Petri Net with Timed Transitions}, reset, inhibit, and query arcs is a tuple $PN^D=(P,T,E,C,W,R,I,Q,QW,Z,D)$, where
$P,T,E,C,W,R,I,Q,QW,Z$ are the same as for $PN^{pri}$, and 
$D : T \rightarrow \mathds{N} \setminus \{0\}$ is a duration function that assigns positive integer durations to transitions.

Figure~\ref{fig:echaintm} shows an extended version of Petri Net model of the Electron Transport Chain~\cite{CampbellBook} shown in Figure~\ref{fig:echain}. The new transitions $tq$ and $tcytc$ (shown in dotted outline) are timed transitions modeling the speed of the small carrier molecules, Coenzyme Q ($q$) and Cytochrome C ($cytc$) as an effect of membrane fluidity. Higher numbers for transition duration represent slower movement of the carrier molecules due to lower fluidity. 
{\bf Execution in $PN^D$} changes, since the token update from $M_k$ to $M_{k+1}$ can involve transitions that started at some time $l$ before time $k$, but finish at $k+1$. Thus, the new marking is computed as follows:
$\forall p \in P \setminus R(T_k), M_{k+1}(p)  = M_k(p)  -  \sum_{\substack{t \in T_k  \wedge p \in \bullet t}} W(p,t) + \sum_{\substack{t \in T_l \wedge p \in t \bullet: l \leq k, l+D(t) = k+1}} W(t,p)$, and 
$\forall p \in R(T_k), \\ M_{k+1}(p) = \sum_{\substack{t \in T_l  \wedge p \in t \bullet : l \leq k, l+D(t) = k+1 }} W(t,p)$, 
where $R(T_i)=\displaystyle\cup_{\substack{t \in T_i}} R(t)$.

A timed transition $t$ produces its output $D(t)$ time units after being fired. We replace $f\ref{f:tparc}$ with $f\ref{f:dur:tparc}$ adding transition duration and replace rule $r\ref{r:add}$ with $r\ref{r:dur:add}$ that produces tokens at the end of transition duration \footnote{\label{fn:enc_dur_nre} We can easily make these timed transitions non-reentrant by adding rule $e\ref{e:ne:dur}$ that disallows a transition from being enabled if it is already in progress:
\vspace{-5pt}
\begin{description}
\item[\nexte:\label{e:ne:dur}] \texttt{\small notenabled(T,TS1):-fires(T,TS0), num(N), TS1>TS0, 
    tparc(T,P,N,C,TS0,D), 
    col(C), time(TS0), time(TS1), TS1<(TS0+D).} %
\end{description}}:
\begin{description}
\item[\nextf:\label{f:dur:tparc}] Rules \texttt{\small tparc($t_i,p_j,n_c,c,ts_k,D(t_i)$):-time($ts_k$).} for each $(t_i,p_j) \in E^+$, $c \in C$, $n_c=m_{W(t_i,p_j)}(c) : n_c > 0$. %
\item[\nextr:\label{r:dur:add}] \texttt{\small add(P,Q,T,C,TSS):-fires(T,TS),time(TS;TSS), tparc(T,P,Q,C,TS,D), \\ TSS=TS+D-1.} %
\end{description}

\begin{proposition}
Let $PN^D$ be a Petri Net with colored tokens, reset, inhibit, read arcs and priority based timed transitions and $M_0$ be an initial marking and let $\Pi^6(PN^D,M_0,k)$ be the ASP encoding of $PN^D$ and $M_0$ over a simulation of length $k$ as defined in Section \ref{sec:enc_dur}. Then $X^6=M_0,T_0,M_1,\dots,T_k$  is an execution sequence of $PN^D$ (with respect to $M_0$) iff there is an answer-set A of $\Pi^6(PN^D,M_0,k)$ such that:
$\{ fires(t,j) : t \in T_j, 0 \leq j \leq k \} = \{ fires(t,ts) : fires(t,ts) \in A \}$, and
$\{ holds(p,q,c,j) : p \in P, c/q \in M_j(p), 0 \leq j \leq k \} \\ = \{ holds(p,q,c,ts) : holds(p,q,c,ts) \in A\}$
\end{proposition}

\section{Example Use of Our Encoding and Reasoning Abilities}

We illustrate the usefulness of our encoding by applying it to the following simulation based reasoning question\footnote{As it appeared in \texttt{https://sites.google.com/site/2nddeepkrchallenge/}} from~\cite{CampbellBook}:
``Membranes must be fluid to function properly. How would decreased fluidity of the membrane affect the efficiency of the electron transport chain?''

To answer this question, first we build a Petri Net model of the Electron Transport Chain, including its interplay with the membrane. Our model is shown in Figure~\ref{fig:echain}. We base our model on~\cite[Figure 9.15]{CampbellBook}, which shows multiple substances flowing through the protein complexes that form this chain. For example, the first complex ($t1$), removes electrons from NADH ($nadh$) arriving at the Mitochondrial Matrix ($mm$) and delivers them to the mobile carrier Coenzyme Q ($q$), converting NADH to NAD+ ($nadp$). As a side effect, $t1$ also moves H+ ($h$) ions from $mm$ to the Intermembrane Space ($is$). The speed at which $q$ (which lives in the membrane) shuttles electrons to next complex depends upon the membrane fluidity.

\begin{figure}
\centering
\vspace{-25pt}
\includegraphics[width=\linewidth]{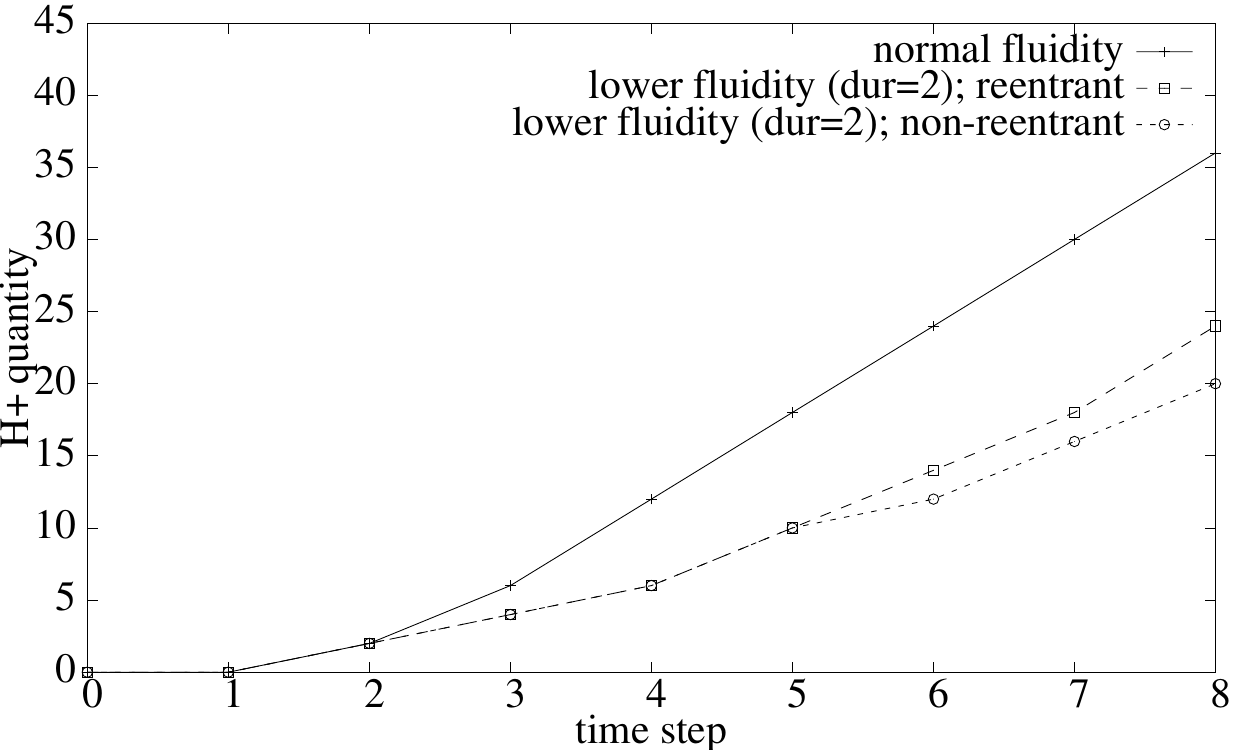}
\vspace{5pt}
\caption{H+ production in the intermembrane space over time for the normal fluidity, lower fluidity (reentrant), and lower fluidity (non-reentrant transitions).}
\vspace{-10pt}
\label{fig:plot_h_prod}
\end{figure}

To answer the question, we need to model change in fluidity and its impact on the mobile carriers. Background knowledge tells us that lower fluidity leads to slower movement of mobile carriers, leading to longer transit times. We model this using an intervention to the Petri Net model of Figure~\ref{fig:echain}, extending it with additional delay transitions in the path of $q$ and Cytochrome C ($cytc$), as shown with dotted outline in Figure~\ref{fig:echaintm}. We encode both models in ASP using encodings from Sections~\ref{sec:enc_cpn} and \ref{sec:enc_dur}, respectively, and simulate them for a fixed number of time-steps $ts$ using the {\em maximal firing set} semantics from Section~\ref{sec:enc_max_firing}. A plot of H+ produced over time is shown in Figure~\ref{fig:plot_h_prod}. We compute the efficiency of the electron transport chain as the quantity of H+ (``$h$'') ions moved (from ``$mm$'') to ``$is$'' over the simulation duration ``$ts$'', i.e. ``$\frac{h}{ts}$''. We found that this value decreased from 4.5 to 3 with decrease of membrane fluidity (modeled as timed transitions of duration 2). Thus, our results show that decreased fluidity of the membrane results in lowering the efficiency of the electron transport chain.

If we permit additional background knowledge about the mobile carriers, we can refine our ASP encoding,  modeling the mobile carriers with non-reentrant timed transitions (Section~\ref{sec:enc_dur} Footnote~\ref{fn:enc_dur_nre}) \footnote{Similar modeling is also possible by using inhibitor arcs from mobile carriers to the transitions preceding them.}. Repeating our simulation with non-reentrant timed transitions  results in an efficiency value of 2.5, which is a larger reduction in the efficiency of the electron chain due to decreased fluidity.

ASP's enumeration of the entire simulation evolution allows us to perform additional reasoning not directly possible with Petri Nets. For example, partial state or firing sequence can be encoded (as ASP constraints) as {\em way-points} to guide the simulation. A simple use case is to enumerate answer-sets where a transition $t$ fires when one of its upstream source products $S$ is found to be depleted. These answer-sets are used to identify another upstream substance responsible for $t$'s firing. 
Our encoding allows various Petri Net dynamic and structural properties  to be easily analyzed, as described in our previous work~\cite{anwar2013pniclp}.

\section{Related Work and Conclusion}
Now we look at some of the existing Petri Net systems that support higher level Petri Net constructs\footnote{The Petri Net Tools Database web-site summarizes a large slice of existing tools \texttt{http://www.informatik.uni-hamburg.de/TGI/PetriNets/tools/quick.html}} (focusing on the ones used for biological modeling). We also look at some ways existing Petri Net tools are used for biological analysis and put them in context of our research.

CPN Tools~\cite{jensen2007coloured}, Renew~\cite{kummer1999renew}, Snoopy~\cite{SnoopyPN} all support Colored Petri Nets. All but CPN Tools directly support inhibit and reset arcs. All but Snoopy are limited to one particular firing semantics, while Snoopy allows three distinct firing semantics. Neither pursues more than one simulation and all break ties arbitrarily. Cell Illustrator~\cite{nagasaki2010cell} does not support colored tokens, but does provide a rich graphical environment with pathway representation similar to standard biological pathways. It also only supports one possible evolution.

Petri Nets have been previously used to analyze biological pathways~\cite{Heiner2006,Hofestadt1998,li2006structural}, but most of this analysis has been limited to dynamic and structural properties of the Petri Net model. \cite{peleg2005using} took a different approach, where they surveyed the Petri Net implementations and came up with questions answerable by each. Contrary to the previous work, we focus on real world biological questions as they appear in college level biological text books; we model these questions as Petri Net extensions; and leverage ASP as a rich reasoning environment.

\textbf{Conclusion:} In this paper we presented the suitability of using Petri Nets with colored tokens for modeling biological pathways. We showed how such Petri Nets can be intuitively encoded in ASP, simulated, and reasoned with, in order to answer real world questions posed in the biological texts. We showed how our initial encoding can be easily extended to include additional extensions, such as maximal firing semantics, priority transitions, and timed transitions. Our encoding has a low specification-implementation gap, it allows enumeration of all possible state evolutions, the ability to guide the search by specifying way-points (such as partial state), and a strong reasoning ability. Our focus in this work is more on encoding flexibility, exploring all possible state evolutions, and reasoning capabilities; and less on performance. We showcased the usefulness of our encoding by an example. We briefly compared our work to other Petri Net systems and their use in biological modeling and analysis. 
In follow on papers, we will extend our work to include the High-level Petri Net standard.

\bibliographystyle{splncs}
\bibliography{cpnasp}

\end{document}